\documentclass[sigconf,10pt,authorversion,nonacm]{acmart}

\usepackage{booktabs} 
\usepackage{color}
\usepackage{url}

\usepackage[ruled]{algorithm2e}

\usepackage[caption=false,font=normalsize,labelfont=sf,textfont=sf]{subfig}

\makeatletter
\def\@ACM@checkaffil{
    \if@ACM@instpresent\else
    \ClassWarningNoLine{\@classname}{No institution present for an affiliation}%
    \fi
    \if@ACM@citypresent\else
    \ClassWarningNoLine{\@classname}{No city present for an affiliation}%
    \fi
    \if@ACM@countrypresent\else
        \ClassWarningNoLine{\@classname}{No country present for an affiliation}%
    \fi
}
\makeatother
\newcommand{\removelatexerror}{\let\@latex@error\@gobble}

\setcopyright{rightsretained}

\acmConference[Mobi'24]{ACM HotMobile conference}{February 28-29, 2024}{San Diego, CA} 
\acmYear{2023}
\copyrightyear{2023}

\acmArticle{4}
\acmPrice{15.00}

\def\name{Confidant}

\begin{document}
\title{Confidant: Customizing Transformer-based LLMs via Collaborative Edge Training}

\author{\large{Yuhao Chen, Yuxuan Yan, Qianqian Yang, Yuanchao Shu, Shibo He and Jiming Chen}}
\affiliation{Zhejiang University}







\renewcommand{\shortauthors}{Y. Chen et al.}

\begin{abstract}
Transformer-based large language models (LLMs) have demonstrated impressive capabilities in a variety of natural language processing (NLP) tasks. Nonetheless, it is challenging to deploy and fine-tune LLMs on mobile edge devices with limited computing, memory, and energy budgets. In this paper, we propose \name, a multi-backend collaborative training framework for customizing state-of-the-art LLMs on commodity mobile devices like smartphones. \name\ partitions an LLM into several sub-models so that each fits into a mobile device's memory. A pipeline parallel training mechanism is further developed to ensure fast and efficient distributed training. In addition, we propose a novel backend scheduler to allocate different attention heads to heterogeneous compute hardware, including mobile CPU and GPUs, to maximize the compute resource utilization on each edge device. Our preliminary experimental results show that \name\ achieves at most 45.3\% memory reduction and 8.03x inference speedup in practical settings.

\end{abstract}

%
%



\maketitle

\section{Introduction}
\label{sec:introduction}
Transformer-based large language models (LLMs), exemplified by models like BERT~\cite{devlin2018bert}, LLaMa~\cite{touvron2023llama2} and GPT-4~\cite{openai2023gpt4}, have ushered in a remarkable era of progress in machine learning and artificial intelligence. Presently, LLM-based applications are predominantly deployed in the cloud, necessitating users to transmit their data to remote servers and await responses. Deploying LLMs on mobile devices allows for the processing of data locally, enhancing data privacy and real-time response. Moreover, compared with cloud-hosted LLMs that provide only general knowledge, locally customized LLMs could learn domain- and user-specific knowledge through fine-tuning, providing much higher inference accuracy and user experience.


\begin{figure}[htbp]
   \centering
   \includegraphics[width=2.7in]{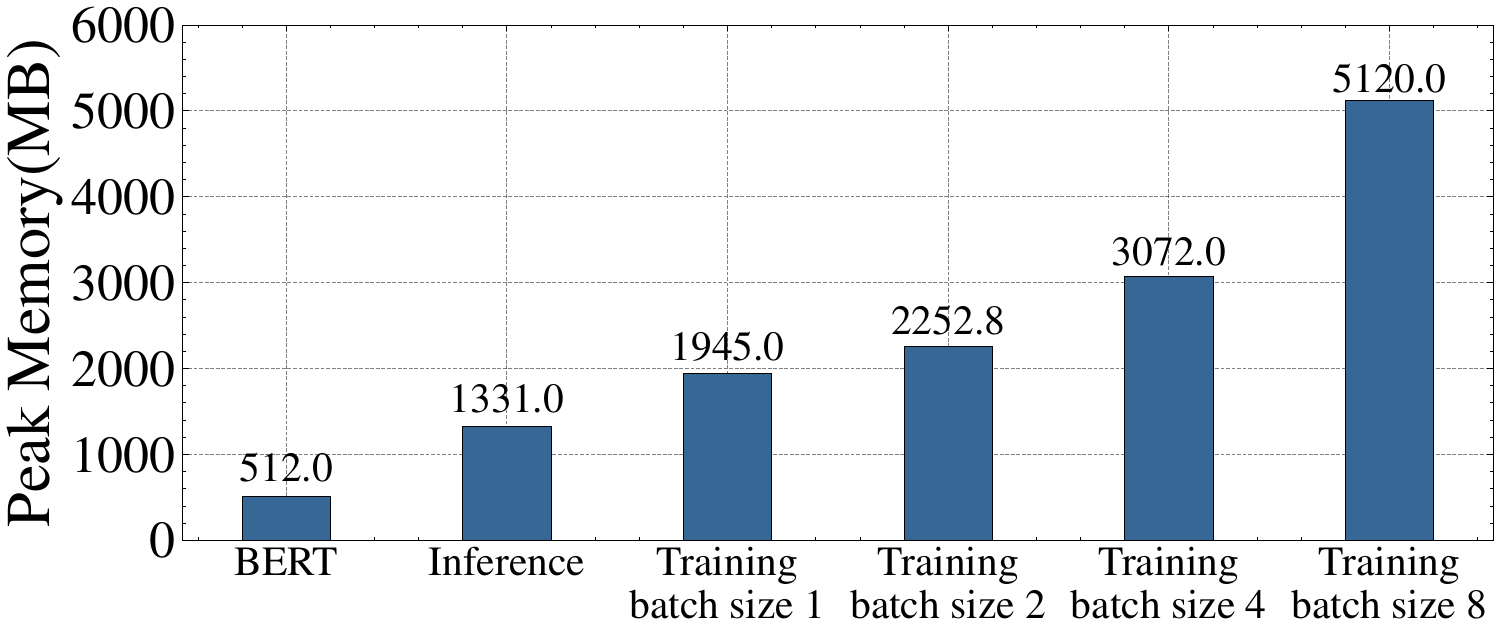}
   \caption{Memory usage of BERT, inference and training with different batch sizes on Redmi K10X Pro using MNN.}
   \label{fig:motivation_mem}
\end{figure}

Fine-tuning an LLM on mobile devices indeed encounters two main challenges. Firstly, LLM training requires way more memory compared to both model inference and the training of CNNs~\cite{chen2016training}. 
To illustrate this, we implemented a BERT model~\cite{devlin2018bert} on a Redmi K10X Pro mobile phone, using MNN~\cite{alibaba2020mnn} and measured the memory usage of the model itself, inference and training with different batch sizes, respectively. As shown in Figure~\ref{fig:motivation_mem}, even fine-tuning the BERT model with a batch size of 8 requires approximately 5GB of memory. Furthermore, we provide a comparison of the fine-tuning memory usage of some state-of-the-art LLMs and the memory capacities of recently released mobile devices in Table~\ref{Tab: Device and llm mem}. While some mobile devices may have memory capacities exceeding 5GB, we note that 5GB of free memory is not always guaranteed, as users may be simultaneously running other applications. These observations highlight the practical limitations of fine-tuning LLMs on resource-constrained mobile devices, emphasizing the need for memory-efficient training techniques.


\begin{table}
\caption{Fine-tuning memory usage of LLMs and memory of the mobile devices released recently.}
\centering
\scalebox{0.75}{
\begin{tabular}{cc|cc}
\hline
LLMs       & Fine-tuning Memory Usage & Mobile Devices           & Memory \\ \hline
Bert-Large & $\sim$16GB             & Huawei Mate 60 Pro       & 12GB    \\ \hline
GPT2-XL   & $\sim$24.5GB                   & Google Pixel 8 Pro       & 12GB    \\ \hline
LLaMA-7B  & $\sim$28GB                   & iPhone 15 Pro                & 8GB    \\ \hline
LLaMA-13B  & $\sim$48GB                 & Galaxy s23 Ultra & 12GB    \\ \hline
\end{tabular}
}
\label{Tab: Device and llm mem}
\end{table}

\begin{figure}
   \centering
   \includegraphics[width=2.7in]{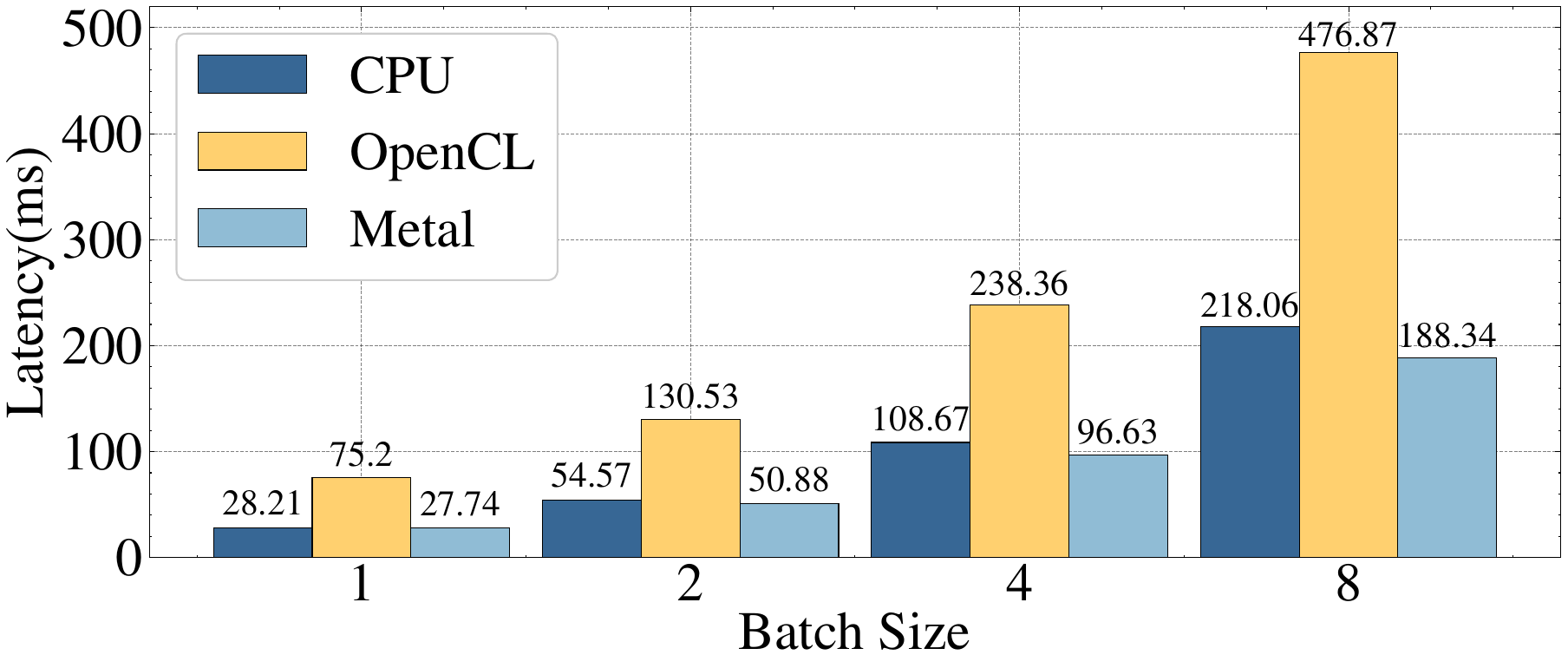}
   \caption{Computation time of the self-attention layer with different batch sizes on an M1 Macbook Pro using different backends.}
   \label{fig:motivation_time}
\end{figure}

Secondly, the limited computing capacities of mobile devices have a direct impact on real-time response and fine-tuning speed. In server environments, current deep neural network (DNN) frameworks generally utilize CUDA, a GPU computation toolkit, as the backend to accelerated computation. The backend, in this context, refers to the underlying engine or computational platform for executing neural network operations. However, mobile devices do not support CUDA on their GPUs. Instead, current industry-scale mobile frameworks typically use alternative backends like OpenCL and Metal to perform DNN computations on mobile GPUs. Despite these efforts, the limited computational capacity of mobile GPUs and the lack of optimization for on-GPU backends in current mobile frameworks results in similar computation latency for DNNs by a CPU and a mobile GPU. To illustrate, we measure the latency of calculating a self-attention layer in BERT with different batch sizes using CPU, OpenCL and Metal, respectively, as the backend on an M1 Macbook Pro laptop, shown in Figure~\ref{fig:motivation_time}. It indicates  that the computation times using these three backends are of the same order of magnitude, and in some cases, using OpenCL is even slower than using CPU. Additionally, it is worth noting that current mobile frameworks only support computing DNNs with a single backend on one device, leaving the available computational capacities not fully utilized.




To tackle the aforementioned challenges, we introduce \textit{Confidant} in this paper, a multi-backend edge collaborative training framework for fine-tuning transformer-based LLMs on mobile devices. Motivated by the fact that individuals often possesses multiple mobile devices, Confidant partitions an LLM into several sub-models, and place them on multiple mobile devices to facilitate collaborative fine-tuning. Confidant follows a pipeline-parallel training mechanism to reduce potential stalling, and further introduces a novel backend scheduler that allocates different numbers of attention heads within a submodel to all available backends on each participating device, accelerating fine-tuning through concurrent training on heterogeneous hardware.  Preliminary results show that fine-tuning an LLM using Confidant achieves up to 45.3\% memory reduction and 8.03x speedup compared to fine-tuning using a single mobile device.
 
Compared with prior work on distributed edge training which targets convolutional-based models, NVIDIA (e.g., Jetson) and non-GPU devices (e.g., Raspberry Pi)~\cite{chen2023ftpipehd, yi2023edgemoe}, to the best of our knowledge, we are the first to implement distributed training of transformer-based LLMs on mobile devices with both CPU and GPU using an industry-scale mobile DNN framework. \name\ further accelerates the training by a novel pipeline-parallel design which leverages unique characteristics of both the Transformer structure and the mobile hardware.

\begin{figure*}[htbp]
   \centering
   \includegraphics[width=5in]{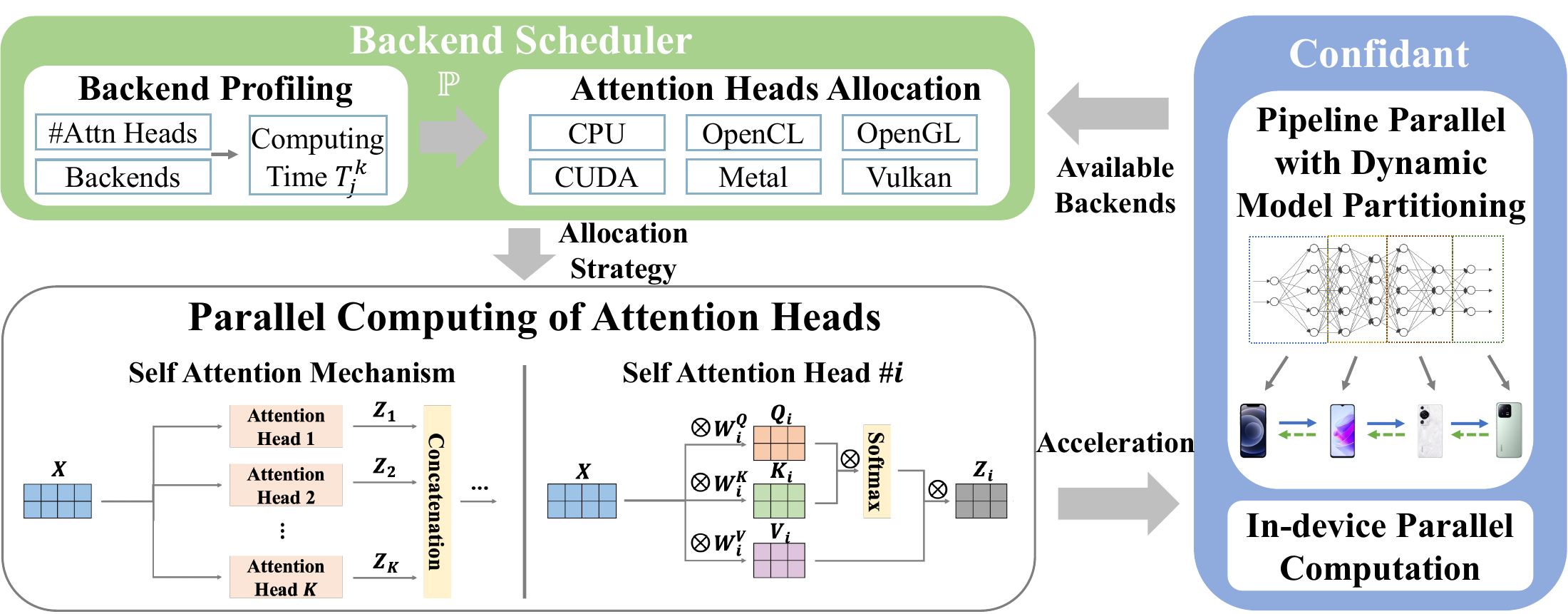}
   \caption{The overview of the proposed system.}
   \label{fig:system_overview}
\end{figure*}

\section{Background}
\subsection{Self Attention Mechanism}
The impressive performance of the LLM is rooted in the transformer structure~\cite{vaswani2017attention}, where the self-attention mechanism plays a pivotal role. The self-attention mechanism typically consists of multiple attention heads which process the input in a parallel manner, as shown in the left-bottom box of Figure~\ref{fig:system_overview}. Each attention head functions as follows: the input \textbf{$X$} is initially transformed by three matrices $W^Q_i$, $W^K_i$ and $W^V_i$, producing the query matrix $Q_i$, key matrix $K_i$ and value matrix $V_i$, respectively. The query matrix $Q_i$ is then multiplied with $K_i$, where the resulting matrix is passed through a softmax function to obtain the attention score matrix. Then the value matrix is multiplied with the attention score matric to obtain the final output of the attention head $Z_i$. Finally, the outputs from all attention heads are concatenated for subsequent processing in the transformer structure. Each attention head with a different set of weights captures distinct patterns within the input data, enabling the transformer structure to capture different dependencies and relationships between elements in the input data.


\begin{figure}[htbp]
   \centering
   \includegraphics[width=3.4in]{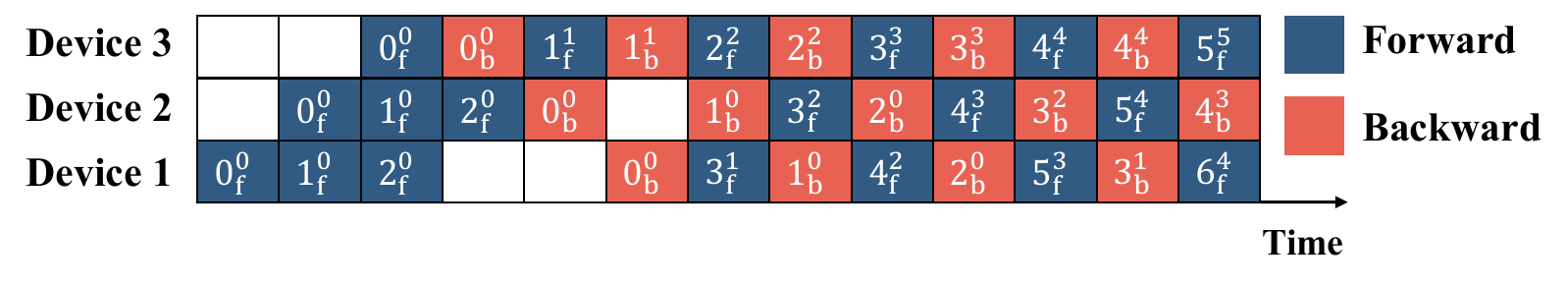}
   \caption{An illustration of the pipeline parallel training technique.}
   \label{fig:pipeline_parallel}
\end{figure}

\subsection{Pipeline Parallel Training}
Pipeline parallel training is one of the widely-used techniques to speed up collaborative training, which was first proposed by Microsoft for training DNN models on GPU clusters~\cite{narayanan2019pipedream}. It first divides a DNN model into several sub-models and trains each sub-model in a different GPU, where the forwarding and the backwarding of a certain batch are executed in a consecutive order. It then accelerates the training by allowing each GPU to forward a batch using stale sub-model weights instead of waiting for the newest sub-model weights. 


We illustrate the idea of pipeline parallel training using three mobile devices as an example in Fig~\ref{fig:pipeline_parallel}, where device 1 is the central node and $i^{ver}_\mathrm{f}$ and $i^{ver}_\mathrm{b}$ denote the forwarding and backwarding of the $i$-th training batch with the weights of the version number $ver$, respectively. Consider the training of batch 1 in device 1, instead of waiting for the backwarding of batch 0, device 1 forwards batch 1 with the same weights used by batch 0. To guarantee the model convergence, the pipeline parallel training applies the one-forward-one-backward (1F1B) rule, guaranteeing edge devices performing the forwarding and backwarding of the local sub-model alternatively. Additionally, the weights utilized to forward a certain batch are stashed for subsequent backwarding of the same batch. In our previously proposed framework FTPipeHD~\cite{chen2023ftpipehd}, we have extended the pipeline parallel training to edge devices with a novel dynamic model partitioning approach. The proposed approach periodically estimates the computing capacity of each worker node and calculates the optimal model partition points accordingly.

\section{Design of Confidant}
\label{sec:design}
\subsection{System Overview}
As shown in Figure~\ref{fig:system_overview}, the proposed system \emph{Confidant} follows the pipeline parallel training mechanism in our previous work, FTPipeHD~\cite{chen2023ftpipehd}, which dynamically partitions the model into several sub-models and trains each of them on a different device in a pipeline parallel manner. 


While FTPipeHD was designed for image models like MobileNetV2 and was implemented on edge devices such as Raspberry Pi, our current work focuses on the distributed training of transformer-based large language models, leading to notable variations in implementation, as discussed in Section~\ref{sec:implementation}. To further accelerate the training process, we extend the approach by utilizing multiple backends within a single device,  enabling parallel computation of attention heads across these backends. We propose a novel backend scheduler that optimally allocates varying numbers of attention heads to all available backends on a device, with the aim of minimizing computation time and improving training efficiency.




\subsection{Backend Scheduler}

The proposed backend scheduler encompasses two key steps: backend profiling and attention head allocation.

\subsubsection{Backend Profiling}
The backend profiling step begins by identifying all available backends on the mobile devices. It then proceeds to profile the computation time of different numbers of attention heads on each of these backends. There are two ways for computing multiple attention heads on a given backend: (1)\textit{Compute them as one large attention head} and (2)\textit{Compute them separately}. In the first way, the $W^Q_i$, $W^K_i$, and $W^V_i$ tensors are concatenated into three larger tensors, $W_Q$, $W_K$, and $W_V$, respectively. This allows for subsequent computations of $Q$, $K$, $V$, and $Z$ to be performed as a single tensor, without the need for separate calculations for each attention head. In the second way, each attention head is calculated individually. The experimental results reveal that, for certain backends, the first way of computing multiple attention heads as a single large attention head is faster than the second method. This performance difference may be attributed to the fact that modern deep neural network (DNN) frameworks incorporate optimizations to accelerate large tensor computations~\cite{jia2022codl, wang2022lightseq2}. For each backend denoted as $j$, we record the time taken to compute $k$ attention heads using both the first and second methods. The faster of the two times is then chosen as the profiling result and is represented as $T_j^k$. These $T_j^k$ values collectively constitute a profiling dataset denoted as $\mathbb{P}$, which is subsequently utilized in the attention head allocation step.

\subsubsection{Attention Heads Allocation}
In this step, the allocation of attention heads is carried out according to Algorithm~\ref{algorithm: allocation}. The objective is to distribute the attention heads to each available backend in such a way that the computation time for the allocated attention heads on each backend is roughly equivalent. More precisely, we employ a binary search approach to identify the minimum execution time required for all the available backends to collectively complete the computation of all the attention heads.


Suppose the goal is to allocate $K$ attention heads to $M$ available backends in an optimized manner. We initialize the binary search by setting the lower bound $l$ to $0$ and the upper bound $r$ to the minimum time required by any backend to compute all $K$ attention heads. Then, iteratively, we calculate the mid-value as $mid = (l+r)/2$. Based on this mid-value, we assess the feasibility of allocating attention heads so that each backend's computation time approximates $mid$. Specifically, for each backend $j$, we identify a number $k$ within the range [$1$, $K$] such that $T_j^k$, time taken by backend $j$ to compute $k$ attention heads, is closest to $mid$. If the deviation of $T_j^k$ from $mid$ exceeds a given threshold $\epsilon$, we assign $k_j=0$. This means we bypass backends that are either too slowly or too fast. If the sum of all assigned $k_j$ values meets or surpasses $K$, an allocation strategy exists where the total computation time for all $K$ attention heads is roughly $mid$. In such a case, we adjust the upper bound for the binary search as $r = mid-\sigma$, where $\sigma$ is a small value to prevent infinite looping. If the sum of all $k_j$ is less than $K$, it suggests that processing all the attention heads within $mid$ time is not feasible. Consequently, the lower bound is adjusted to $l = mid - \sigma$. The binary search ends when $l > r$. Finally, based on the determined allocation vector $\mathbb{S}$, attention heads are distributed among the device's backends. These backends then execute their allocated computations in parallel during training, enhancing the efficiency of the process.

    \begin{algorithm}[h]
        \scriptsize
        \caption{Allocation of attention heads on multiple backends}
        \label{algorithm: allocation}
        \LinesNumbered 
        \label{feature encoding algorithm}
        \KwIn{The profiling set $\mathbb{P}$, total attention heads $K$, total available backends $M$, threshold $\epsilon$.}
        \KwOut{The allocation strategy $\mathbb{S} = \{(j,k_j) | j=1,...,M, k_j =0, 1,...,K\}. $}

        Initialize $l \leftarrow 0$, $r \leftarrow $ $\min\limits_{j = 1,...M} T_j^K$,  $\mathbb{S} \leftarrow \{\}$;
        
        \While{$l \le r$}{
            $mid \leftarrow (l+r)/2$;

            \eIf{$isValid(mid, K, \mathbb{P})$}{
                $r \leftarrow mid - \sigma$;
            }{
                $l \leftarrow mid + \sigma$;
            }
        }

        return $\mathbb{S}$;

        \BlankLine
        
        \SetKwFunction{validFunc}{$isValid$}
        \SetKwProg{Fn}{Function}{:}{}
        \Fn{\validFunc{$mid, K, \mathbb{P}$}}{
            Initialize $totalHeads \leftarrow 0$, $\mathbb{S}' \leftarrow \{\}$;
            
            \For{$j \leftarrow 1$ \KwTo $M$}{
                $k_j=\arg\min\limits_{k=1, ... K} abs(T_j^k - mid)$;

                \If{$abs(T_j^{k_j} - mid) > \epsilon$}{
                    $k_j \leftarrow 0$;     
                }
                
                $totalHeads \leftarrow totalHeads + k_j$;

                insert $(j, k_j)$ into $S'$;
            }

            \eIf{$totalHeads >= K$}{
                $\mathbb{S} \leftarrow \mathbb{S}'$;

                return $true$;
            }{
                return $false$;
            }
        }
    \end{algorithm}
    

\subsection{Implementation}
\label{sec:implementation}
We then develop an application on mobile devices for the implementation of Confidant by developing based on MNN~\cite{alibaba2020mnn} of version 2.7.0 as the deep learning framework. MNN supports training DNN models on Android and iOS and outperforms other frameworks in terms of computation time and memory usage as shown in \cite{wang2022melon}. Nonetheless, given that MNN is implemented in C++ while the application is developed in Java, we employ the Java Native Interface (JNI)~\cite{enwiki:1166267665} to invoke C++ functions from within the application. The communication between mobile devices is facilitated using standard HTTP requests.

Our work is pioneering in implementing the transformer model for training on MNN. The process of loading pre-trained weights of a LLM from the corresponding PyTorch model involves the following steps: PyTorch-formatted pre-trained weights are first converted into the ONNX format, which is an open standard for representing machine learning models; The MNNConvert tool provided by MNN is then employed to load the weights in ONNX format into the MNN framework.

To enhance MNN's functionality for our specific requirements, certain modifications are made to its source code. These modifications include extending MNN's support for passing tensors to the $step(x)$ function, which initially only supports scalars. In pipeline parallel training, the device may need to call $step(x)$ with $x$ being gradient tensors, thus necessitating this extension. To parallelly compute multiple attention heads on multiple backends, we create separate computation graphs for each backend. In these graphs, we allocate different numbers of attention heads decided by the backend scheduler. This ensures efficient parallelization of attention head computations across the backends, ultimately contributing to the overall acceleration of the training process.


\section{Evaluation}
\label{sec:evaluation}
\subsection{Evaluation Settings}
In our preliminary evaluations, we leverage BERT for a classification task using the Conll2003 dataset~\cite{sang2003introduction}. We first load the pre-trained BERT weights from Pytorch following the procedure outlined in Section~\ref{sec:implementation}. Then we append a fully connected layer to the output of BERT for classification purposes. The pretrained BERT model, along with the added fully connected layer, is collectively fine-tuned using the proposed Confidant across multiple mobile devices, with their specifications detailed in Table~\ref{Tab: Device List}.



\subsection{Memory Usage}
\begin{figure}[htbp]
   \centering
   \includegraphics[width=3.35in]{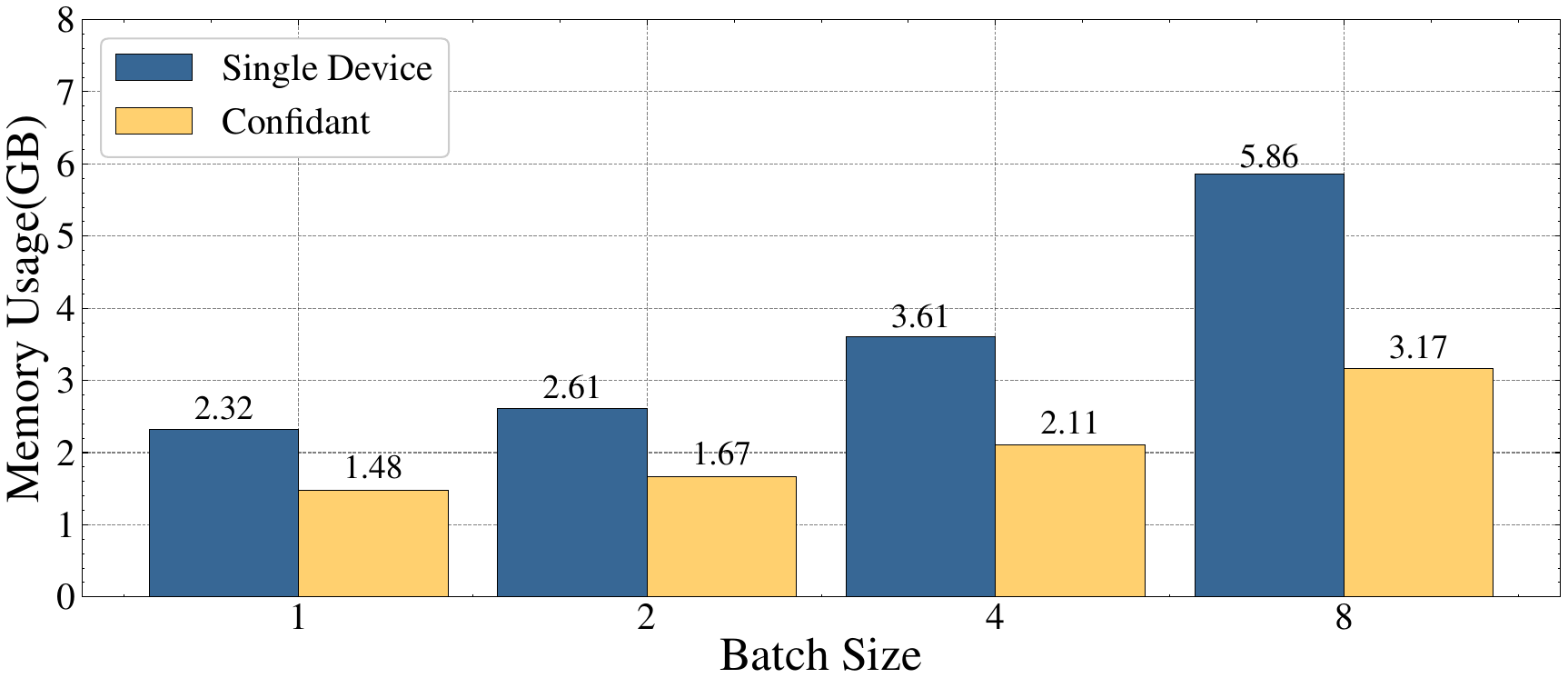}
   \caption{Memory usage comparison on BERT fine-tuning between Confidant and a single device.}
   \label{fig:evaluation_mem}
\end{figure}

We compute the average memory usage across the listed three phones in Table~\ref{Tab: Device List} during training with Confidant and compare it to the memory usage when training with a single device, Redmi 10X Pro, as illustrated in Figure~\ref{fig:evaluation_mem}. We can observe a substantial reduction in average memory usage when employing Confidant compared to training on a single device. Note that while Confidant utilizes three mobile phones for fine-tuning, its memory usage doesn't scale linearly. This is because each phone stores multiple versions of weights and intermediate outputs of the sub-model to ensure compliance with the 1F1B (One Forward, One Backward) rule for pipeline parallel training. Nonetheless, Confidant manages to achieve a notable memory reduction of 45.3\% when employing a batch size of 8.


\subsection{Multi-backend Parallel}
\begin{figure}[htbp]
   \centering
   \includegraphics[width=3.35in]{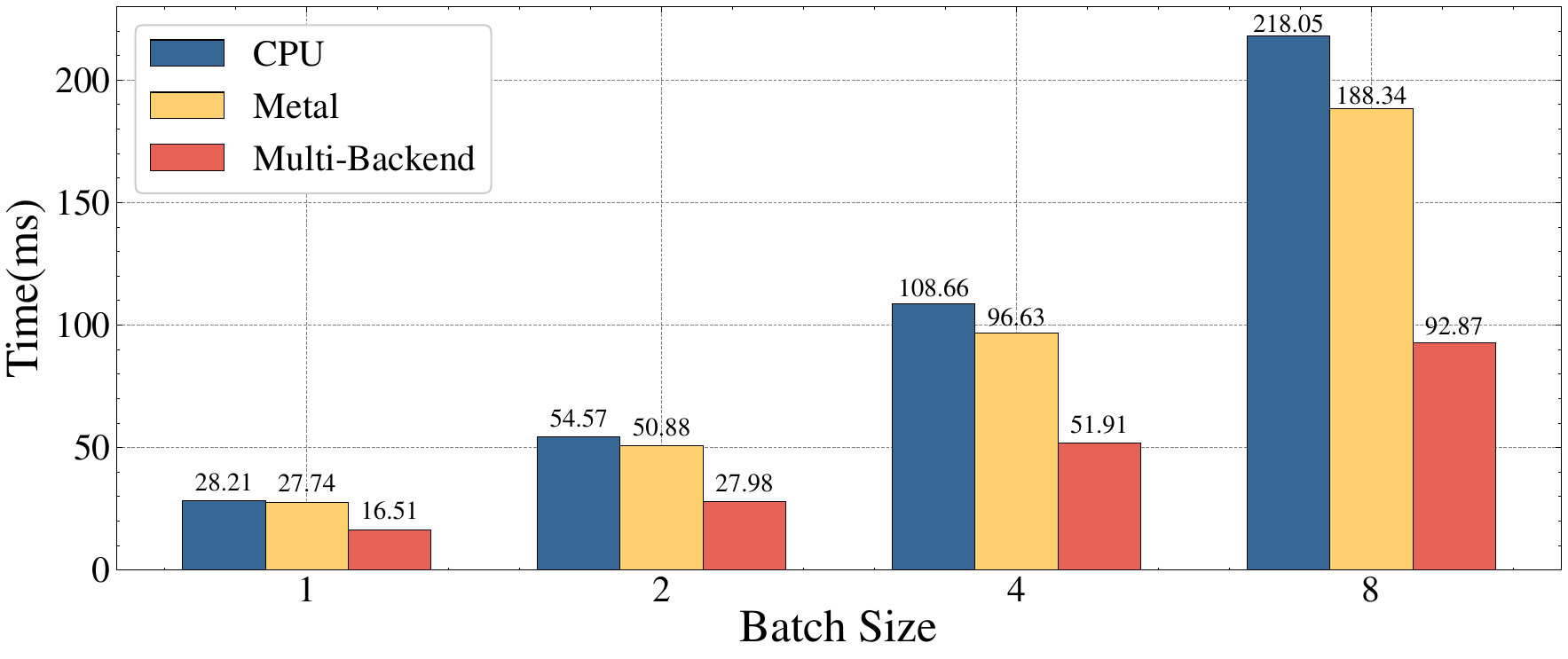}
   \caption{Computation time of a self-attention layer using only CPU, only Metal, and both.}
   \label{fig:evaluation_multi_backend}
\end{figure}

Next, we evaluate the performance of the proposed multi-backend parallelism. We execute the backend scheduler on a Macbook Pro, utilizing the available backends, namely CPU, OpenCL, and Metal, to  collaboratively compute a self-attention layer with 12 attention heads. The backend scheduler selects CPU and Metal while omitting OpenCL due to its slower computation speed, as previously demonstrated in Figure~\ref{fig:motivation_time}. Figure~\ref{fig:evaluation_multi_backend} illustrates the latency of single-backend computations using either CPU or Metal, alongside the latency of multi-backend computations that utilize both CPU and Metal. It is clearly shown that multi-backend computations offer a significant acceleration compared to using a single backend.

 

\begin{table}
\caption{Device List}
\centering
\scalebox{0.75}{
\begin{tabular}{ccccc}
\hline
Device & Redmi 10X Pro & Redmi K50      & Mi 10 Lite      & MBP \\ \hline
Soc    & Dimensity 820 & Dimensity 8100 & Snapdragon 765G & M1    \\ \hline
GPU    & Mali-G57 MC5  & Mali-G610      & Adreno 620      & M1    \\ \hline
Memory & 8G            & 12G            & 8G              & 16G         \\ \hline
\end{tabular}
}
\label{Tab: Device List}
\end{table}

\begin{figure*}[htbp]
\centering
	\subfloat[]{
    \label{fig:evaluation_training_1}
      \includegraphics[height=1.1in,width=0.25\linewidth]{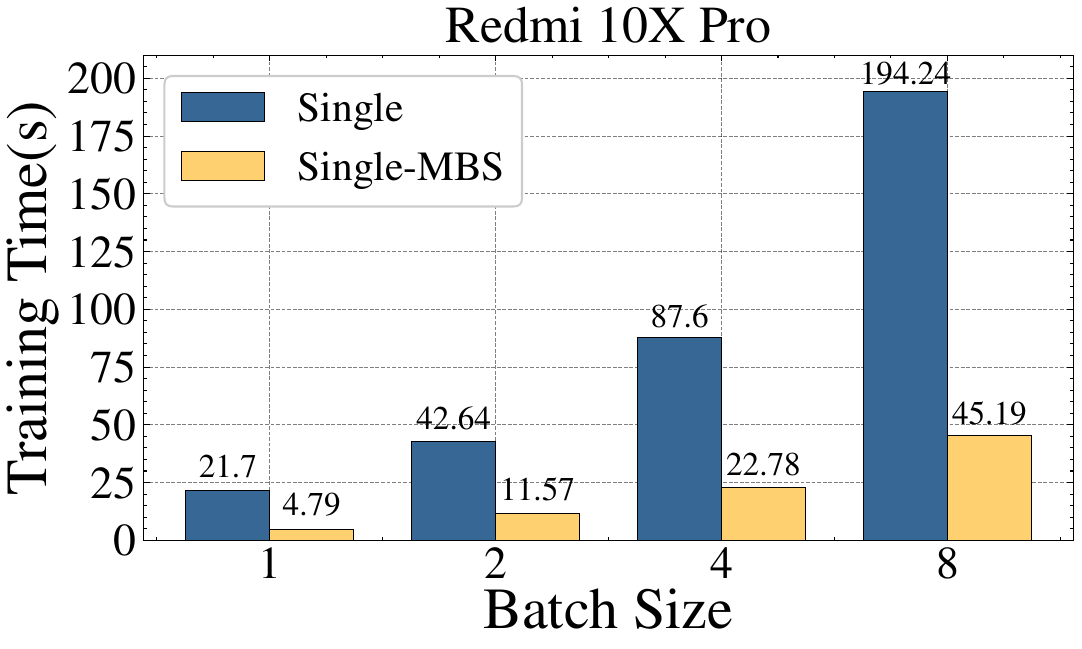}}
    \subfloat[]{
    \label{fig:evaluation_training_2}
      \includegraphics[height=1.1in,width=0.25\linewidth]{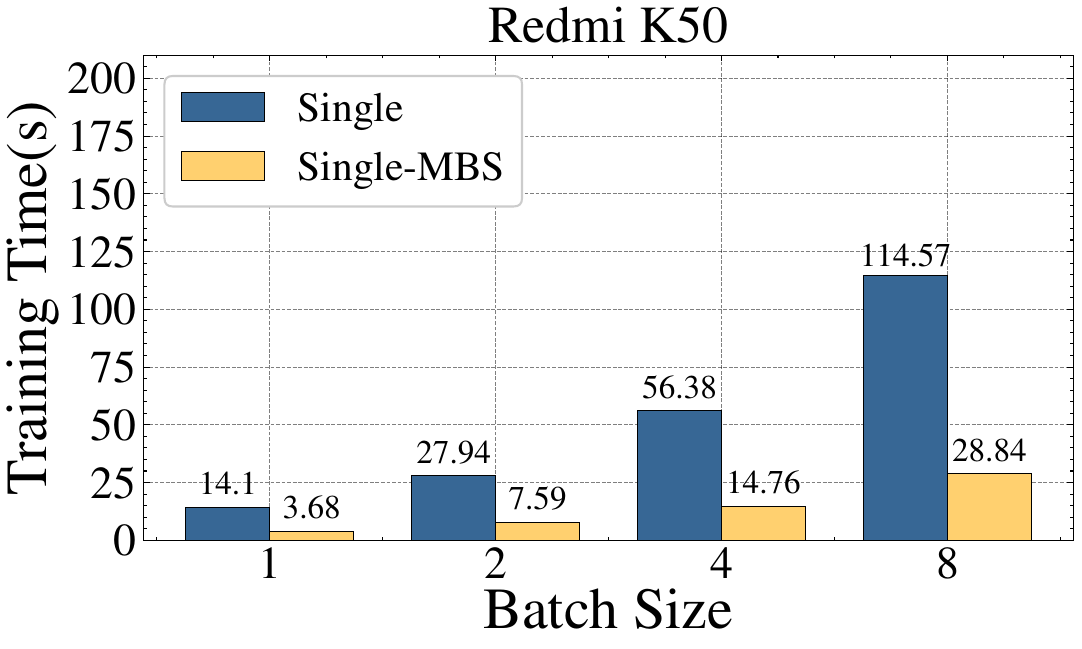}}
    \subfloat[]{
    \label{fig:evaluation_training_3}
     \includegraphics[height=1.1in,width=0.25\linewidth]{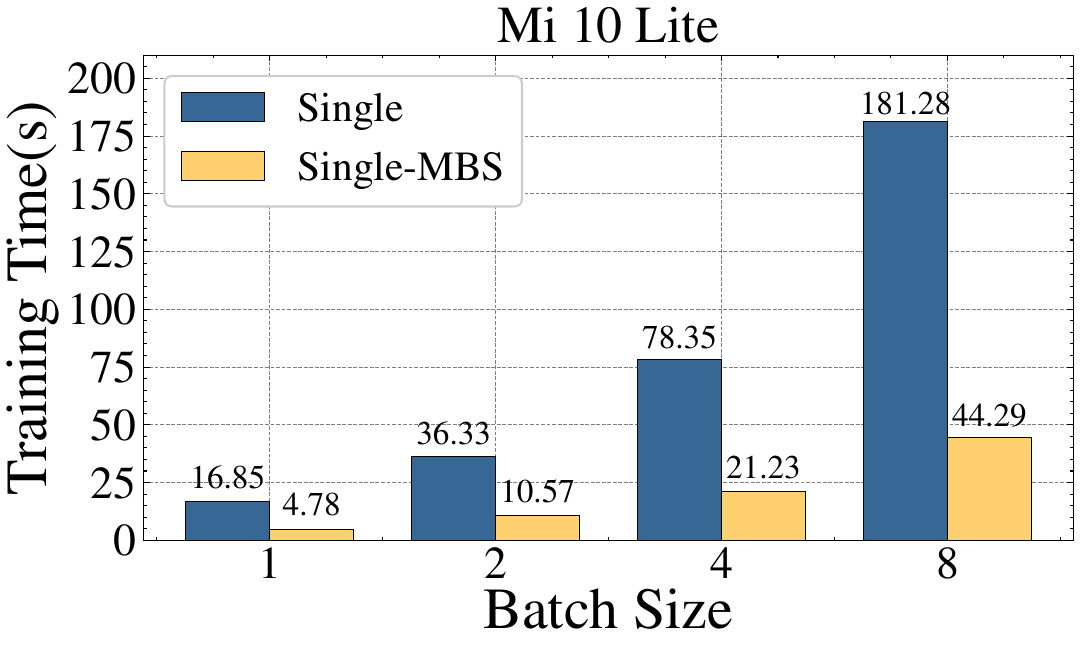}}
   \subfloat[]{
   \label{fig:evaluation_training_4}
     \includegraphics[height=1.1in,width=0.25\linewidth]{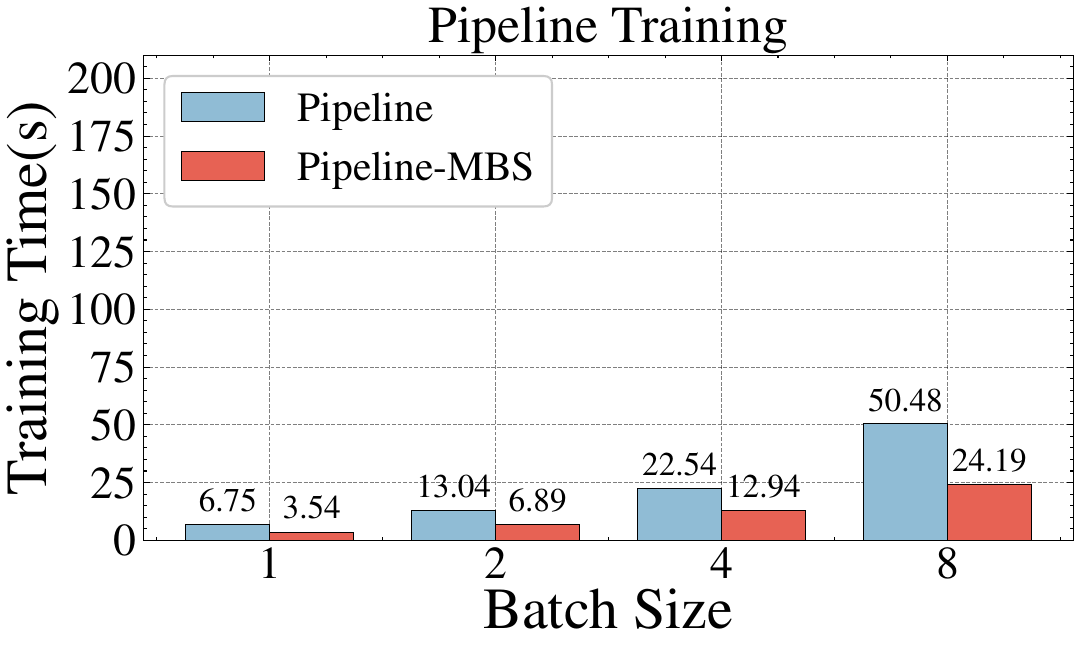}}
\captionsetup{justification=centering}
\caption{Training latency comparisons between single devices and our system}
\label{fig:evaluation_training_comparison}
\end{figure*}

\subsection{Training Latency}
Finally, we analyze the training latency by Confidant under different batch sizes, which encompasses the total time required for both the forward and backward passes of a single batch. We compare it against three baselines here, i.e., training on a single device with a single backend, referred to as \emph{Single}, training on a single device with mutiple backends utilizing the proposed backend scheduler, referred to as \emph{Single-MBS}, pipeline parallel training across three devices using one backend on each, referred to as \emph{Pipeline}.

We conduct a comparison of \emph{Single} and \emph{Single-MBS} using three different mobile phones individually to demonstrate the performance of the proposed multiple backend scheduler. The results are presented in Figure~\ref{fig:evaluation_training_1}, \ref{fig:evaluation_training_2} and \ref{fig:evaluation_training_3}. From the results, it is evident that by implementing the multiple backend scheduler, significant speedups of up to 4.53x, 3.97x, and 4.09x are achieved on the three respective phones.

We then compare the training latency by Confidant with that by \emph{Pipeline}, as shown Figure~\ref{fig:evaluation_training_4}. The experimental results clearly demonstrate that Confidant achieves a maximum speedup of 2.09 times compared to pipeline parallel training. This outcome further confirms the effectiveness of the proposed multiple backend scheduler.

Comparing Figure~\ref{fig:evaluation_training_4} with Figure~\ref{fig:evaluation_training_1}, \ref{fig:evaluation_training_2} and \ref{fig:evaluation_training_3}, we can observe that the pipeline parallel training achieves a maximum speedup of 3.84x compared to training on a single phone. This validates the advantage of  pipeline parallel training in accelerating the training process. Furthermore, it's worth noting that the proposed Confidant achieves remarkable speedup ratios of up to 8.03x compared to training on a single phone, underlining its significant performance gains.

\section{Discussion and Future Work}
\label{sec:discussion}
This paper summarizes our progress and preliminary experimental results on collaborative edge training of LLMs. We intend to further explore and investigate the following aspects.

\textbf{Memory Adaptation.} Given that users may concurrently run applications during LLM training on a mobile device, the available memory for training can vary over time and even abruptly change, leading to training interruptions caused by insufficient memory. In our forthcoming research, we plan to develop a memory adaption technique that can effectively adapt the training process to varying memory budgets.

\textbf{Fault Tolerance for Energy-aware Training.} 
Given that many mobile devices operate on battery power, an energy-aware training algorithm could effectively improve the reliability of the fine-tuning process. We also aim to improve the reliability by designing a proper fault tolerance mechanism. This strategy involves predicting the remaining battery life and proactively relocating training workloads to devices with sufficient battery power. 

\textbf{Cross Framework Implementation.} As previously discussed in Section~\ref{sec:introduction}, individuals may possess edge devices of different types, each of which supports different deep neural network (DNN) frameworks (e.g., MNN for mobile phones and Pytorch for laptops). As such, it becomes imperative to develop methods to collaboratively train a large language model (LLM) using diverse mobile DNN frameworks.

\section{Conclusion}
\label{sec:conclusion}

In this paper, we introduced Confidant, a multi-backend edge collaborative training framework designed for the fine-tuning of transformer-based large language models (LLMs) on mobile devices. Our approach involves partitioning the LLM into multiple sub-models and distributing them across several mobile devices. We leveraged pipeline parallel training and dynamic model partitioning to expedite the training process. Additionally, we introduced a novel backend scheduler to further enhance training speed, dynamically allocating varying numbers of attention heads to multiple backends and enabling parallel computation across multiple backends within one device. We implemented Confidant on mobile devices using an industry-scale deep neural network (DNN) platform, and conducted preliminary evaluations to demonstrate its efficacy in reducing memory usage on individual devices while simultaneously accelerating the training process.

\bibliographystyle{ACM-Reference-Format}
\bibliography{ref} 

\end{document}